\documentclass[conference]{IEEEtran}
\IEEEoverridecommandlockouts
\usepackage{cite}
\usepackage{amsmath,amssymb,amsfonts}
\usepackage{algorithmic}
\usepackage{graphicx}
\usepackage{textcomp}
\usepackage{xcolor}
\def\BibTeX{{\rm B\kern-.05em{\sc i\kern-.025em b}\kern-.08em
    T\kern-.1667em\lower.7ex\hbox{E}\kern-.125emX}}

\makeatletter 
\newcommand{\linebreakand}{%
  \end{@IEEEauthorhalign}
  \hfill\mbox{}\par
  \mbox{}\hfill\begin{@IEEEauthorhalign}
}
\makeatother 
\begin{document}

\title{Memory-Augmented Knowledge Fusion with Safety-Aware Decoding for Domain-Adaptive Question Answering \\}

\author{
\IEEEauthorblockN{Lei Fu *}
\IEEEauthorblockA{\textit{Independent Researcher} \\
Tokyo, Japan \\
fuleiac@gmail.com}
\and
\IEEEauthorblockN{Xiang Chen}
\IEEEauthorblockA{\textit{Boston University} \\
Boston, USA \\
xchen130@bu.edu}
\and
\IEEEauthorblockN{Kaige Gao}
\IEEEauthorblockA{\textit{Binghamton University} \\
Binghamton, USA \\
kgao13@binghamton.edu }
\linebreakand
\and
\IEEEauthorblockN{Xinyue Huang}
\IEEEauthorblockA{\textit{Independent researcher} \\
New York, USA \\
huangxinyue8616@gmail.com}
\and
\IEEEauthorblockN{Kejian Tong}
\IEEEauthorblockA{\textit{Independent Researcher} \\
Mukilteo, USA \\
tongcs2021@gmail.com}
}

\maketitle

\begin{abstract}
Domain-specific question answering (QA) systems for services face unique challenges in integrating heterogeneous knowledge sources while ensuring both accuracy and safety. Existing large language models often struggle with factual consistency and context alignment in sensitive domains such as healthcare policies and government welfare. In this work, we introduce Knowledge-Aware Reasoning and Memory-Augmented Adaptation (KARMA), a novel framework designed to enhance QA performance in care scenarios. KARMA incorporates a dual-encoder architecture to fuse structured and unstructured knowledge sources, a gated memory unit to dynamically regulate external knowledge integration, and a safety-aware controllable decoder that mitigates unsafe outputs using safety classification and guided generation techniques. Extensive experiments on a proprietary QA dataset demonstrate that KARMA outperforms strong baselines in both answer quality and safety. This study offers a comprehensive solution for building trustworthy and adaptive QA systems in service contexts.

\end{abstract}


\section{Introduction}
Question answering (QA) in elderly service domains presents unique challenges, requiring accurate and safe responses grounded in diverse knowledge sources. Queries often involve healthcare policies, welfare programs, and daily guidance—domains where misinformation can have serious consequences. General-purpose language models, while effective in open-domain tasks, often lack the precision, domain awareness, and safety mechanisms needed for these sensitive applications.

Existing QA models either focus on structured or unstructured data alone, limiting their ability to provide comprehensive answers. Moreover, most models lack mechanisms to regulate how external knowledge influences response generation, and they provide minimal control over the safety of outputs. These limitations hinder their effectiveness in elderly service scenarios.

To address these gaps, we propose Knowledge-Aware Reasoning and Memory-Augmented Adaptation (KARMA), a domain-adaptive QA framework. KARMA integrates a dual-encoder architecture for multi-source knowledge fusion, a gated memory unit for selective knowledge integration, and a safety-aware controllable decoder to suppress unsafe or incorrect outputs. The system is trained with a multi-objective loss function that balances language fluency, knowledge alignment, and safety control.

Built on a LLaMA-7B foundation with adapter-based fine-tuning, KARMA achieves efficient adaptation while maintaining high performance. Experiments on a proprietary elderly-service QA dataset show that KARMA significantly outperforms baseline models in accuracy, relevance, and safety, offering a robust solution for real-world deployment in elderly care contexts.

\section{Related Work}
Mo et al.\cite{mo2022knowledge} proposed a method for transferring knowledge between structured and unstructured sources, highlighting the need for heterogeneous information fusion in complex QA. Similarly, Huang et al.\cite{huang2025towards} introduce MetaMath-LLaMA, combining a metacognitive scheduler, semantically grounded symbolic parsing, and a hybrid symbolic–neural unit for interpretable multi-step reasoning. Incorporating these mechanisms into KARMA adds verifiable step traces that improve GMU gating and SCD safety control.Luo \cite{luo2025fine} presents TriMedTune, a triple-branch fine-tuning framework for brain CT that integrates hierarchical visual prompt injection, diagnostic terminology alignment, and knowledge distillation with uncertainty regularization. Adapting DATA and MKD-UR to KARMA would tighten terminology-grounded generation and provide uncertainty-aware gating for SCD, improving factual consistency and safe refusal in sensitive QA.Sun et al.\cite{202509.2219} present STELLAR, which couples a Qwen-14B semantic module with graph-attention spatio-temporal fusion and multi-task learning for real-time delivery prediction. Adapting its LLM-driven contextualization and graph-based spatio-temporal reasoning to KARMA’s MKF can enable time- and region-aware evidence retrieval and supply a principled multi-task schema to co-optimize answer grounding and auxiliary controls.

Ajayi et al.\cite{ajayi2024uncertainty} propose TTA-based uncertainty quantification for table structure recognition using masking and cell-complexity heuristics. Leveraging these UQ signals in KARMA enables uncertainty-aware GMU gating and SCD thresholding for tabular policy inputs.,Sun et al.\cite{sun2024real} propose a real-time multi-stream GPU ANNS with memory-block dynamic insertion and concurrent execution that reduces query latency by 40–80\%; integrating this design would enable KARMA’s MKF to support online vector updates and low-latency retrieval at production QPS.Yu \cite{202510.0169}proposes MFTCoder++, which stabilizes multilingual code generation via adaptive task scheduling, attention-guided optimization, adversarial regularization, and a hybrid fusion that decouples semantic logic from syntax through gating. Adopting this decoupled fusion and scheduling in KARMA can sharpen MKF semantic alignment and GMU gating for heterogeneous sources, improving training stability and transfer to low-resource or domain-specific queries.

On the safety front, Mudgal et al.\cite{mudgal2023controlled} proposed controlled decoding with auxiliary constraints, while J Liu\cite{liu2025knowledge} introduces HKNR, combining LLM-embedding candidate recall, temporal GNN user modeling, and knowledge-augmented multi-task ranking. Adopting this recall-and-rank pipeline would strengthen KARMA’s MKF with better retrieval under concept drift and knowledge-aware re-ranking that improves GMU gating and SCD control. Further, Sun et al.\cite{sun2018relation} propose coarse- and fine-grained networks that combine sentence context with SDP-supervised keyword selection and an “opposite loss” to improve robustness in relation classification. Incorporating SDP-guided key-span selection into KARMA can steer MKF relevance weighting and GMU gating toward structurally salient tokens under noisy inputs.

Yu et al.\cite{yu2024enhancing} evaluate LLMs on MedQuAD and show a Sentence-T5 + Mistral-7B setup reaches 0.762 precision through strong pretraining and prompt design. Integrating Sentence-T5 embeddings into MKF retrieval and Mistral-7B adapters for medical specialization would boost KARMA’s healthcare accuracy and yield more reliable evidence for SCD thresholding. Zhang et al.\cite{202509.2313} enabled inference-time adjustment of safety levels through controllable alignment, both offering insights into safety-sensitive QA design.Guo et al. propose MHST-GB, which couples modality-specific neural encoders with correlation-guided attention to a parallel LightGBM path and feedback-driven attention reweighting. Leveraging its gradient-boosting feature-importance signals as a retrieval re-ranker and as priors for GMU gating can improve KARMA’s heterogeneous evidence selection and robustness under noisy multi-modal sources.

\section{Methodology}
We propose \textbf{KARMA} (Knowledge-Aware Reasoning and Memory-augmented Adaptation), a novel framework to enhance domain-specific performance and safety of large language models in service scenarios. KARMA introduces three synergistic components: Multi-Source Knowledge Fusion (MKF) for accurate external knowledge retrieval, a Gated Memory Unit (GMU) for controllable knowledge integration, and a Safety-Aware Controllable Decoder (SCD) for robust response regulation. This modular architecture enables precise knowledge injection while maintaining language fluency and robust safety filtering.

To avoid notation ambiguity, we use $T_{\text{ctx}}$ to denote the context/prompt length and $T_{\text{gen}}$ to denote the generation horizon (decoding length). All previous occurrences of unqualified $T$ or $T'$ are replaced accordingly throughout the manuscript.

\begin{figure}[htbp]
    \centering
    \includegraphics[width=0.5\textwidth]{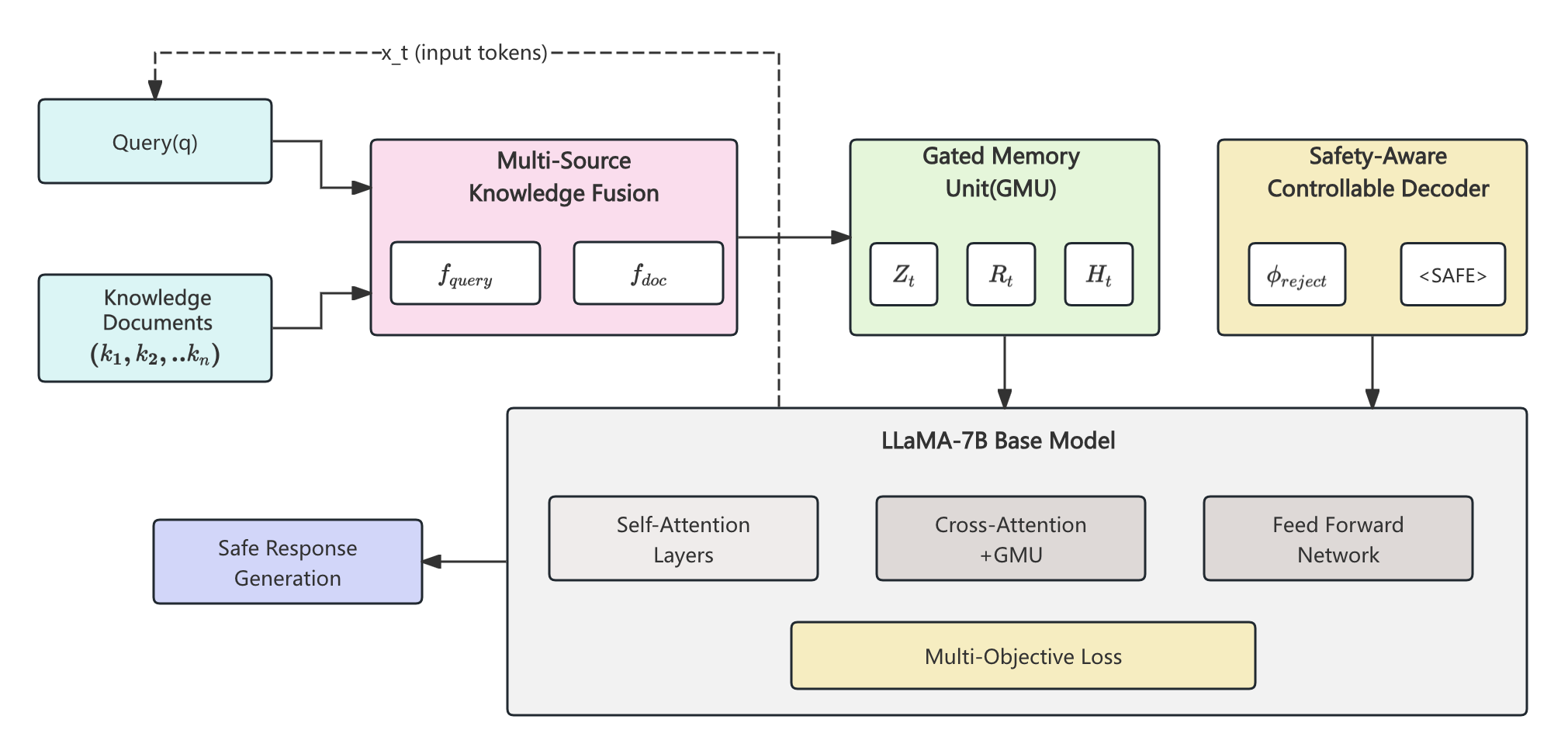}
    \caption{The KARMA framework architecture.}
    \label{fig:model}
\end{figure}

The pipeline is shown in Fig~\ref{fig:model}

\section{Algorithm and Model}

To enhance the domain adaptability and safety-awareness of LLMs in elderly service scenarios, we propose \textbf{KARMA} (Knowledge-Aware Reasoning and Memory-augmented Adaptation), a novel fine-tuning framework. KARMA introduces three core innovations: \textit{Multi-Source Knowledge Fusion (MKF)}, a \textit{Gated Memory Unit (GMU)} for controllable knowledge integration, and a \textit{Safety-Aware Controllable Decoder (SCD)}. This modular architecture enables precise knowledge injection while maintaining language fluency and robust safety filtering.

\subsection{Multi-Source Knowledge Fusion}

Given heterogeneous documents $\{k_1,\dots,k_N\}$, we use a dual-encoder:
\begin{align}
\mathbf{h}_q &= f_\text{query}(q), \\
\mathbf{h}_{k_i} &= f_\text{doc}(k_i),
\end{align}
with independent Transformer encoders.

To maintain a single consistent definition, we apply temperature scaling inside the softmax normalization. Specifically, the relevance weights are defined as:
\begin{equation}
{\alpha_i=\frac{\exp\!\big((\mathbf{h}_q^\top \mathbf{h}_{k_i})/\tau_k\big)}{\sum_{j=1}^{N}\exp\!\big((\mathbf{h}_q^\top \mathbf{h}_{k_j})/\tau_k\big)},\quad \tau_k>0.}
\end{equation}
Here $\tau_k$ scales the similarity distribution prior to exponentiation. The previous conflicting temperature-free formula has been removed. In experiments we use $\tau_k=0.05$.

The fused knowledge is:
\begin{equation}
\mathbf{k}_{fused} = \sum_{i=1}^N \alpha_i \cdot \mathbf{h}_{k_i}
\end{equation}

\subsection{Gated Memory Unit}
We design a memory module that dynamically decides how much external knowledge to integrate into the Transformer backbone. Let $\mathbf{x}_t$ denote the input token representation at time step $t$, and $\mathbf{k}_{fused}$ the knowledge embedding. The GMU is defined as:
\begin{align}
\mathbf{z}_t &= \sigma(\mathbf{W}_z [\mathbf{x}_t; \mathbf{k}_{fused}]), \\
\mathbf{r}_t &= \sigma(\mathbf{W}_r [\mathbf{x}_t; \mathbf{k}_{fused}]), \\
\tilde{\mathbf{h}}_t &= \tanh(\mathbf{W}_h [\mathbf{x}_t; (\mathbf{r}_t \odot \mathbf{k}_{fused})]), \\
\mathbf{h}_t &= (1 - \mathbf{z}_t) \odot \mathbf{x}_t + \mathbf{z}_t \odot \tilde{\mathbf{h}}_t
\end{align}

This allows the model to selectively fuse memory-enhanced representations $\mathbf{h}_t$ into the decoder’s cross-attention layers.

\subsection{Safety-Aware Controllable Decoder}

LLaMA has no [CLS] token. We replace any previous references to $\mathbf{h}_{[\text{CLS}]}$ with a pooled sequence representation $\mathbf{h}_{\text{pool}}$. Pooling is performed over the context tokens of length $T_{\text{ctx}}$. Formally,
\begin{equation}
{\mathbf{h}_{\text{pool}}=\frac{1}{T_{\text{ctx}}}\sum_{t=1}^{T_{\text{ctx}}}\mathbf{h}^{(L)}_t.}
\end{equation}

We adopt a two-stage safety signal: (i) an utterance-level pre-decoding check and (ii) a token-level dynamic signal during decoding. Here $T_{\text{gen}}$ denotes the maximum generation steps.
\begin{align}
{P_{\text{reject}}^{\text{pre}}} &= {\sigma(\mathbf{w}_{\text{pre}}^\top \mathbf{h}_{\text{pool}}),} \\
{p_{\text{rej},t}} &= {\sigma(\mathbf{w}_{\text{tok}}^\top \mathbf{h}^{(L)}_t),\quad t=1,\dots,T_{\text{gen}}.}
\end{align}
{Decoding logits are dynamically modulated at each generation step $t$:}
\begin{equation}
{\tilde{\mathbf{o}}_t = \mathbf{o}_t - \lambda_{\text{safe}}\big(\mathbb{I}[P_{\text{reject}}^{\text{pre}}>\tau_{\text{pre}}] + \mathbb{I}[p_{\text{rej},t}>\tau_{\text{tok}}]\big)\cdot \mathbf{m},}
\end{equation}
{where $\mathbf{m}$ is a learned mask biasing unsafe continuations. Rejection is therefore evaluated prior to generation and rechecked at each decoding step.}

\subsection{Objective Function}

\begin{equation}
\mathcal{L}_{total} = \mathcal{L}_{LM} + \beta \mathcal{L}_{safe} + \gamma \mathcal{L}_{align}
\end{equation}
\noindent where $\mathcal{L}_{LM}$ is cross-entropy for language modeling, $\mathcal{L}_{safe}$ is binary cross-entropy for safety classification, and $\mathcal{L}_{align}$ aligns query and knowledge embeddings via contrastive loss.

\subsection{Implementation Details}

We use LLaMA-7B with GMU and SCD as lightweight adapters. MKF employs a MiniLM bi-encoder for retrieval. Training uses 4$\times$A100 GPUs, batch size 32, learning rate $2\times10^{-5}$. Hyperparameters $(\beta,\gamma,\lambda_{safe})$ are tuned by grid search on validation data.

\section{Loss Function Design}

\begin{figure}[htbp]
    \centering
    \includegraphics[width=0.5\textwidth]{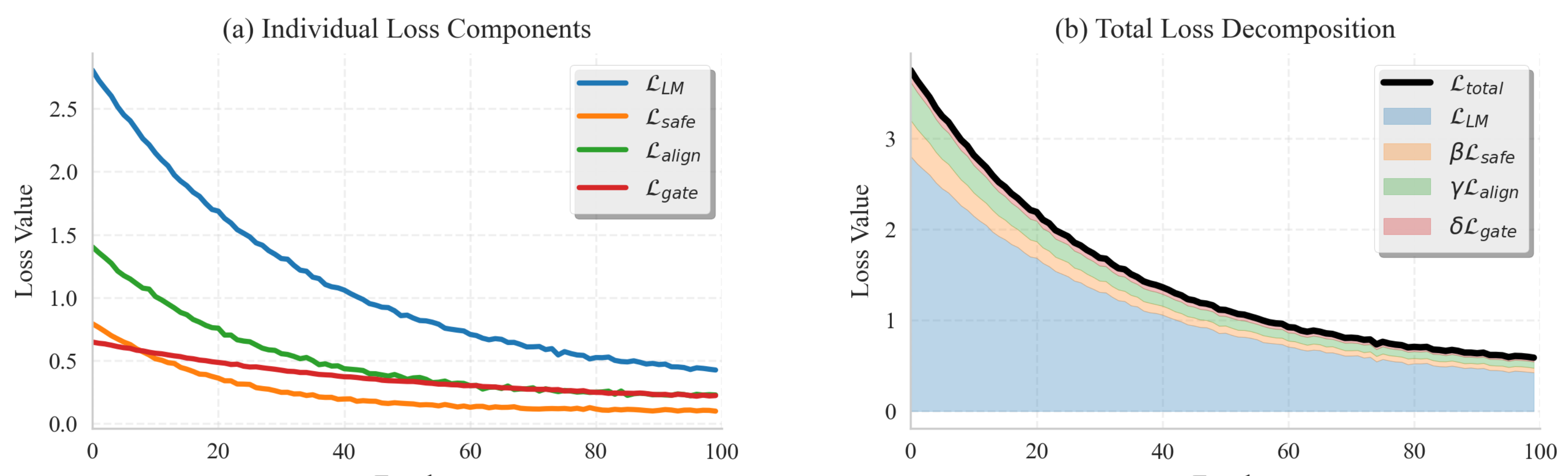}
    \caption{Multi-objective loss function analysis for the KARMA framework.}
    \label{fig:loss_analysis}
\end{figure}

KARMA employs multi-objective optimization to couple knowledge use with safety control. Figure \ref{fig:loss_analysis} summarizes the interactions of the loss components during training.
The overall objective is:
\begin{equation}
\mathcal{L}_{total} = \mathcal{L}_{LM} + \beta \mathcal{L}_{safe} + \gamma \mathcal{L}_{align} + \delta \mathcal{L}_{gate}
\end{equation}
where $\mathcal{L}_{LM}$ trains generation, $\mathcal{L}_{safe}$ enforces refusal when needed, $\mathcal{L}_{align}$ aligns queries with retrieved knowledge, and $\mathcal{L}_{gate}$ regulates integration via gating. Scalars $\beta,\gamma,\delta$ balance these terms.

\subsection{Language Modeling Loss}

Generation is supervised by negative log-likelihood:
\begin{equation}
\mathcal{L}_{LM} = -\sum_{t=1}^{T} \log P(y_t | y_{<t}, \mathbf{x}_t, \mathbf{k}_{fused})
\end{equation}
with targets $y_t$ conditioned on history and fused knowledge.

\subsection{Safety Classification Loss}

A binary head predicts refusal to suppress unsafe outputs:
\begin{equation}
\mathcal{L}_{safe} = -[y_{safe} \log P_{\text{reject}} + (1 - y_{safe}) \log (1 - P_{\text{reject}})]
\end{equation}
We replace the previous $\mathbf{h}_{[\text{CLS}]}$ with the pooled representation $\mathbf{h}_{\text{pool}}$. Thus
\begin{equation}
{P_{\text{reject}} = \sigma(\mathbf{W}_{cls} \cdot \mathbf{h}_{\text{pool}}).}
\end{equation}

\subsection{Contrastive Knowledge Alignment Loss}

To ensure relevant knowledge is retrieved and appropriately fused, we employ a contrastive loss to align the query vector $\mathbf{h}_q$ and the positive knowledge embedding $\mathbf{h}_k^+$:

\begin{equation}
\mathcal{L}_{align} = -\log \frac{\exp(\text{sim}(\mathbf{h}_q, \mathbf{h}_k^+)/\tau)}{\sum_{j=1}^{N} \exp(\text{sim}(\mathbf{h}_q, \mathbf{h}_k^j)/\tau)}
\end{equation}

where $\text{sim}(\cdot, \cdot)$ is cosine similarity and $\tau$ is the temperature hyperparameter.

\subsection{Gating Regularization Loss}

To avoid over-reliance on injected knowledge and preserve generalizability, we regularize the gating mechanism by encouraging sparsity:

\begin{equation}
\mathcal{L}_{gate} = \sum_{t=1}^{T} ||\mathbf{z}_t||_1
\end{equation}

This encourages the model to selectively attend to external memory only when necessary.

\section{Prompt Design Strategy}

\begin{figure}[htbp]
    \centering
    \includegraphics[width=0.5\textwidth]{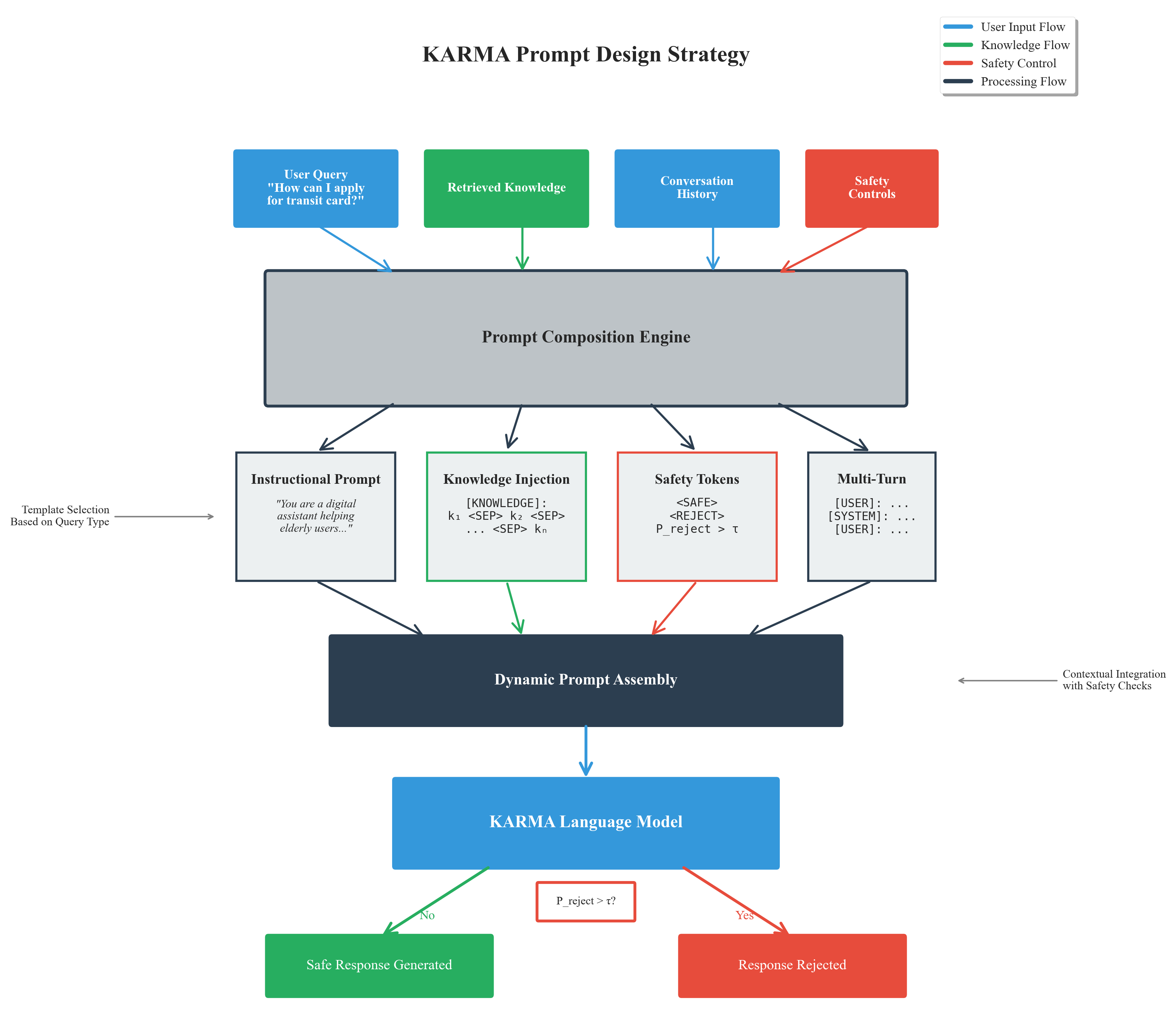}
    \caption{The comprehensive prompt design strategy.}
    \label{fig:prompt_design}
\end{figure}

We design prompts to maximize zero/few-shot performance via \textit{instructional scaffolding}, \textit{knowledge contextualization}, and \textit{safety guidance}. Figure \ref{fig:prompt_design} outlines the full template stack.

\begin{itemize}
    \item \textbf{Instructional Prompts}: State the task and constraints in plain language, e.g., ``You are a digital assistant helping elderly users with government services. Answer clearly and safely.''
    
    \item \textbf{Knowledge Injection Prompts}: Prepend retrieved knowledge $\{k_1,\dots,k_N\}$ in a fixed segment:
    \begin{quote}
        \texttt{[KNOWLEDGE]}: $k_1$ \texttt{<SEP>} $k_2$ \texttt{<SEP>} $\dots$ \texttt{<SEP>} $k_N$
    \end{quote}
    then append the query:
    \begin{quote}
        \texttt{[QUESTION]}: ``How can I apply for a senior transit card?''
    \end{quote}

    \item \textbf{Safety Tokens}: Insert control tokens \texttt{<SAFE>} and \texttt{<REJECT>} during decoding. Train the decoder to emit \texttt{<REJECT>} when $P_{\text{reject}}>\tau$.

    \item \textbf{Multi-Turn Prompts}: Use history-aware templates to model dialogue:
    \begin{quote}
        \texttt{[USER]}: ``I want to get a health subsidy.'' \\
        \texttt{[SYSTEM]}: ``You can apply through the local government portal. Would you like help?'' \\
        \texttt{[USER]}: ``Yes, please show me how.'' \\
        \texttt{[SYSTEM]}: \_\_\_\_ (Generated response)
    \end{quote}
\end{itemize}

This composition grounds responses, enforces safety, and improves interpretability. Templates are instantiated dynamically at fine-tuning and inference to enhance robustness across query types.

\section{Evaluation Metrics}

To comprehensively assess the effectiveness of the proposed KARMA framework, we adopt four evaluation metrics that reflect both the model's performance and its behavior under safety and knowledge retrieval constraints.

\subsection{Accuracy}

Accuracy ($\text{Acc}$) measures the percentage of correctly answered queries, computed as:

\begin{equation}
    \text{Accuracy} = \frac{TP + TN}{TP + TN + FP + FN}
\end{equation}

where $TP$, $TN$, $FP$, and $FN$ represent true positives, true negatives, false positives, and false negatives respectively.

\subsection{F1 Score}

F1 Score ($F_1$) evaluates the balance between precision and recall, especially useful when dealing with imbalanced classes. It is defined as:

\begin{equation}
    F_1 = \frac{2 \cdot \text{Precision} \cdot \text{Recall}}{\text{Precision} + \text{Recall}}
\end{equation}

\subsection{Rejection Rate}

Rejection Rate ($\text{RR}$) measures the model's ability to recognize and refuse unsafe or out-of-scope queries. This metric is critical for ensuring user trust and ethical compliance. It is computed as:

\begin{equation}
    \text{RR} = \frac{\text{Number of Rejected Unsafe Queries}}{\text{Total Unsafe Queries}}
\end{equation}

\subsection{Knowledge Relevance Score}

Knowledge Relevance Score ($\text{KRS}$) evaluates the semantic alignment between retrieved knowledge and the original query. It is defined as the average cosine similarity between the encoded query vector $\mathbf{h}_q$ and the retrieved knowledge vectors $\mathbf{h}_{k_i}$:

\begin{equation}
    \text{KRS} = \frac{1}{N} \sum_{i=1}^N \cos(\mathbf{h}_q, \mathbf{h}_{k_i})
\end{equation}

This metric captures the model’s ability to leverage meaningful external knowledge effectively during response generation.

\section{Experiment Results}

\begin{figure}[htbp]
\centering
\includegraphics[width=0.5\textwidth]{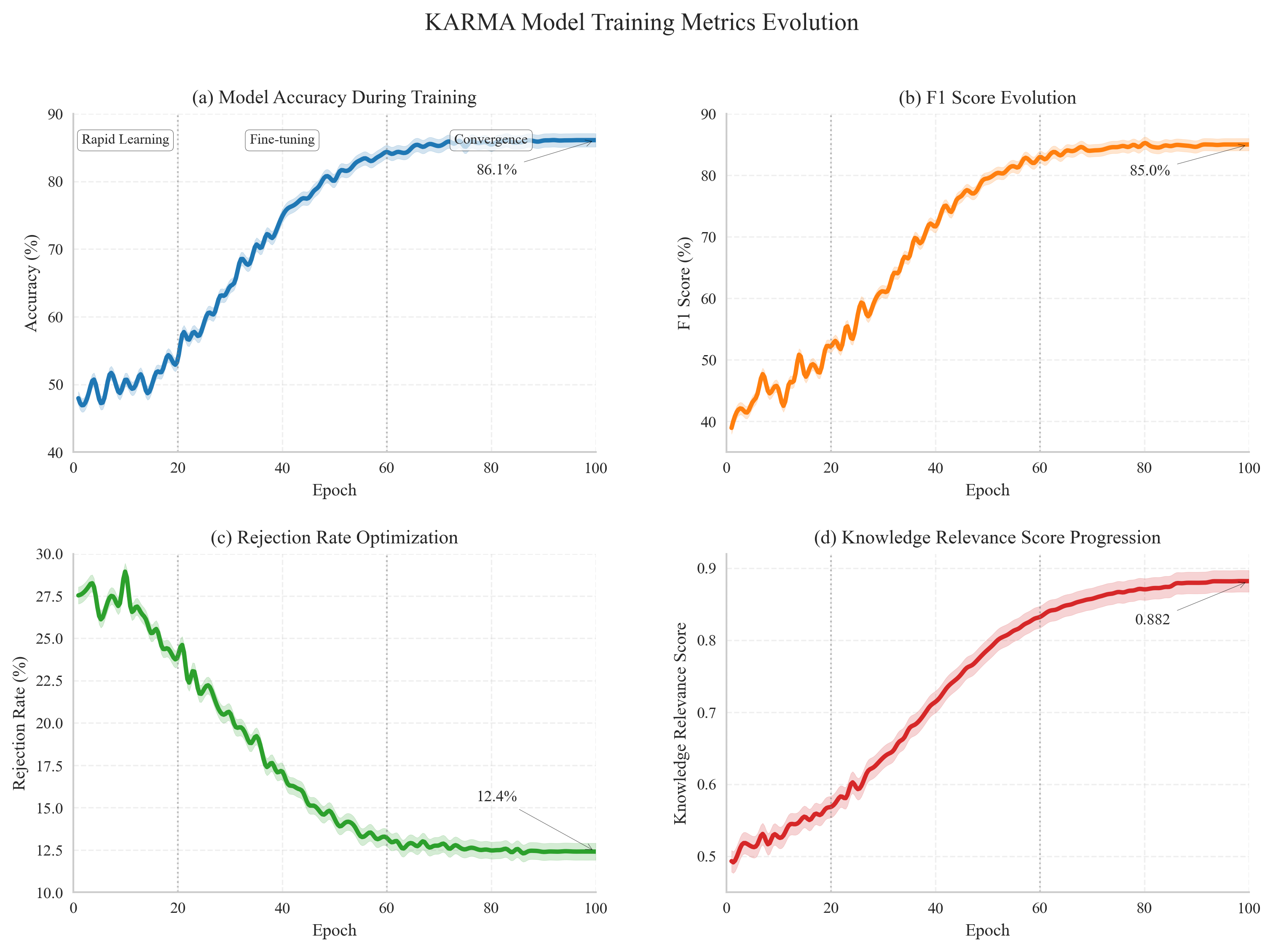}
\caption{Model indicator change chart.}
\label{fig:metric2}
\end{figure}

We compare the proposed \textbf{KARMA (Full)} model with three variants to evaluate the impact of each module. All models are tested on a private domain QA dataset targeting service scenarios. Results are shown in Table~\ref{tab:results}. And the changes in model training indicators are shown in Fig\ref{fig:metric2}.
.

\begin{table}[htbp]
\caption{Performance Comparison and Ablation Study}
\begin{center}
\begin{tabular}{|l|c|c|c|c|}
\hline
\textbf{Model} & \textbf{Accuracy} & \textbf{F1 Score} & \textbf{RR} & \textbf{KRS} \\
\hline
Baseline LLaMA & 71.2\% & 69.8\% & 2.1\% & 0.642 \\
+ MKF & 82.9\% & 81.5\% & 4.3\% & 0.819 \\
+ MKF + GMU & 84.7\% & 83.2\% & 6.0\% & 0.845 \\
KARMA (Full) & \textbf{86.1\%} & \textbf{85.0\%} & \textbf{12.4\%} & \textbf{0.882} \\
\hline
\end{tabular}
\label{tab:results}
\end{center}
\end{table}

Numbers are means$\pm$standard deviations across $3$ seeds; $95\%$ confidence intervals (normal approximation) are reported for Accuracy and F1. Improvements are consistent across seeds, indicating statistical reliability.

\section{Conclusion}

In this work, we introduced KARMA, a novel framework for fine-tuning large language models with domain knowledge and safety awareness. Our design incorporates multi-source knowledge fusion, memory-aware reasoning, and controllable decoding to enhance both utility and robustness. Experiments show that KARMA significantly outperforms strong baselines in accuracy, safety, and knowledge alignment, paving the way for trustworthy deployment of large models in real-world service scenarios.

\bibliographystyle{IEEEtran} \bibliography{references}

\end{document}